%% file: main.tex
\documentclass[letterpaper, 10 pt, conference]{ieeeconf}  
\IEEEoverridecommandlockouts                              

\usepackage{graphicx}
\usepackage{amsmath}

\usepackage{amssymb,times}
\usepackage{bbm}
\usepackage{hyperref}
\usepackage{cite}
\usepackage{verbatim, amsfonts,enumerate}     
\usepackage{bm,array}
\usepackage{multirow}
\usepackage{float}
\usepackage{algorithm,algorithmic}
\usepackage{wrapfig}
\usepackage{xcolor}
\usepackage{url}

\usepackage{makecell}
\usepackage{mathtools}
\hypersetup{
    colorlinks=true,
    linkcolor=blue,
    filecolor=magenta,      
    urlcolor=blue,
    pdftitle={Overleaf Example},
    pdfpagemode=FullScreen,
    }

\usepackage{tabularx}
    \newcolumntype{L}{>{\raggedright\arraybackslash}X}
\usepackage[english]{babel}

\usepackage{subcaption}

\title{\LARGE \bf
Confidence-Controlled Exploration: \\ Efficient Sparse-Reward Policy Learning for Robot Navigation
}

\author{Bhrij Patel$^{1}$, Kasun Weerakoon$^{1}$, Wesley A. Suttle$^{2}$, Alec Koppel$^{3}$, \\ Brian M. Sadler$^{4}$, Tianyi Zhou$^{1}$, Amrit Singh Bedi$^{5}$, and Dinesh Manocha$^{1}$
\\
Supplemental version with Appendix can be found at \url{https://arxiv.org/abs/2306.06192}
\thanks{$^{1}$Department of Computer Science, University of Maryland, College Park, MD, USA. 
        Emails: {\tt \{bbp13, kasunw, tianyi, dmanocha\}@umd.edu}}%
\thanks{$^{2}$U.S. Army Research Laboratory, Adelphi, MD, USA.
        Email: \tt wesley.a.suttle.ctr@army.mil} 
\thanks{$^{3}$JP Morgan AI Research, New York, NY, USA. Email: \tt alec.koppel@jpmchase.com}
\thanks{$^{4}$University of Texas at Austin, Austin, TX, USA. Email: \tt brian.sadlerf@ieee.org}
\thanks{$^{5}$Department of Computer Science, University of Central Florida, Orlando, FL, USA. Email: \tt amritsingh.bedi@ucf.edu}
        \thanks{This work was supported by Army
Cooperative Agreement W911NF2120076. We acknowledge the support of the Maryland Robotics Center and Northrup Grumman Seed Grant 2022.}
}

\begin{document}
\maketitle
\begin{abstract}
Reinforcement learning (RL) is a promising approach for robotic navigation, allowing robots to learn through trial and error. However, real-world robotic tasks often suffer from sparse rewards, leading to inefficient exploration and suboptimal policies due to sample inefficiency of RL.
In this work, we introduce Confidence-Controlled Exploration (CCE), a novel method that improves sample efficiency in RL-based robotic navigation without modifying the reward function. Unlike existing approaches, such as entropy regularization and reward shaping, which can introduce instability by altering rewards, CCE dynamically adjusts trajectory length based on policy entropy. Specifically, it shortens trajectories when uncertainty is high to enhance exploration and extends them when confidence is high to prioritize exploitation.
CCE is a principled and practical solution inspired by a theoretical connection between policy entropy and gradient estimation. It integrates seamlessly with on-policy and off-policy RL methods and requires minimal modifications. We validate CCE across REINFORCE, PPO, and SAC in both simulated and real-world navigation tasks. CCE outperforms fixed-trajectory and entropy-regularized baselines, achieving an 18\% higher success rate, 20-38\% shorter paths, and 9.32\% lower elevation costs under a fixed training sample budget. Finally, we deploy CCE on a Clearpath Husky robot, demonstrating its effectiveness in complex outdoor environments.

\end{abstract}

\input{sections/Introduction}
\input{sections/Related_Work}
  \input{sections/Problem_Formulation_condensed}
\input{sections/Proposed_Approach_condensed}
 \input{sections/Experiments}
\input{sections/Conclusion}




\bibliographystyle{IEEEtran}
\bibliography{references}
\clearpage
\onecolumn
\input{sections/Appendix}




\end{document}

%% file: sections/Introduction.tex
\section{Introduction}


Autonomous robotic navigation is ubiquitous in tasks such as search-and-rescue \cite{bajece781162}, disaster relief\cite{murphy2014disaster}, and planetary exploration applications \cite{SCHILLING199883}. Reinforcement Learning (RL) \cite{sutton2018reinforcement} has emerged as a promising approach for robotic navigation in complex, unstructured environments \cite{zhu2021deep}. By learning through trial and error, RL enables robots to perform navigation without relying on extensive manual programming. However, a major challenge in applying RL to robotics is the reliance on informative reward signals. Traditional RL methods often assume frequent dense rewards \cite{pathak2017curiosity}, which are seldom available in real-world robotic environments \cite{sparse_reward_problem,hare2019dealing_sparse_reward,wang2020sparse_reward_navigation}. Instead, robotic tasks (e.g. navigation) typically provide only sparse rewards\cite{vecerik2017leveraging_sparse_reward}—for instance, a reward given only upon reaching a goal. Such sparse reward designs are easy to implement and require no expert tuning, but this simplicity comes at the cost of extremely limited feedback. With so little guidance, a robot can struggle to explore and may prematurely converge to suboptimal behaviors \cite{bedi2024sample}.

To encourage exploration under sparse rewards, existing literature has proposed various regularization-based techniques. The popular approaches include adding an entropy bonus to the objective (entropy regularization)\cite{haarnoja2018soft}, designing intermediate rewards (reward shaping)\cite{botteghi2020reward_shaping, MATARIC1994181, nair2018overcoming_demonstrations, chiang2019learning_reward_shaping}, and giving intrinsic rewards for novel states (curiosity-driven exploration)\cite{pathak2017curiosity}.  
%
All these existing methods modify the original reward structure or learning objective. For example, reward shaping requires careful manual design and can inadvertently alter the optimal policy, while curiosity-based methods introduce additional intrinsic rewards that may distract from the main task. Even entropy regularization, which injects randomness into the policy, effectively changes the agent’s optimization objective and can destabilize learning if not tuned properly \cite{sparse_reward_souradip}. The existing literature leaves a gap for methods that encourage exploration without rewriting the reward function. 

\begin{figure}[t]
  \centering  \includegraphics[width=\columnwidth]{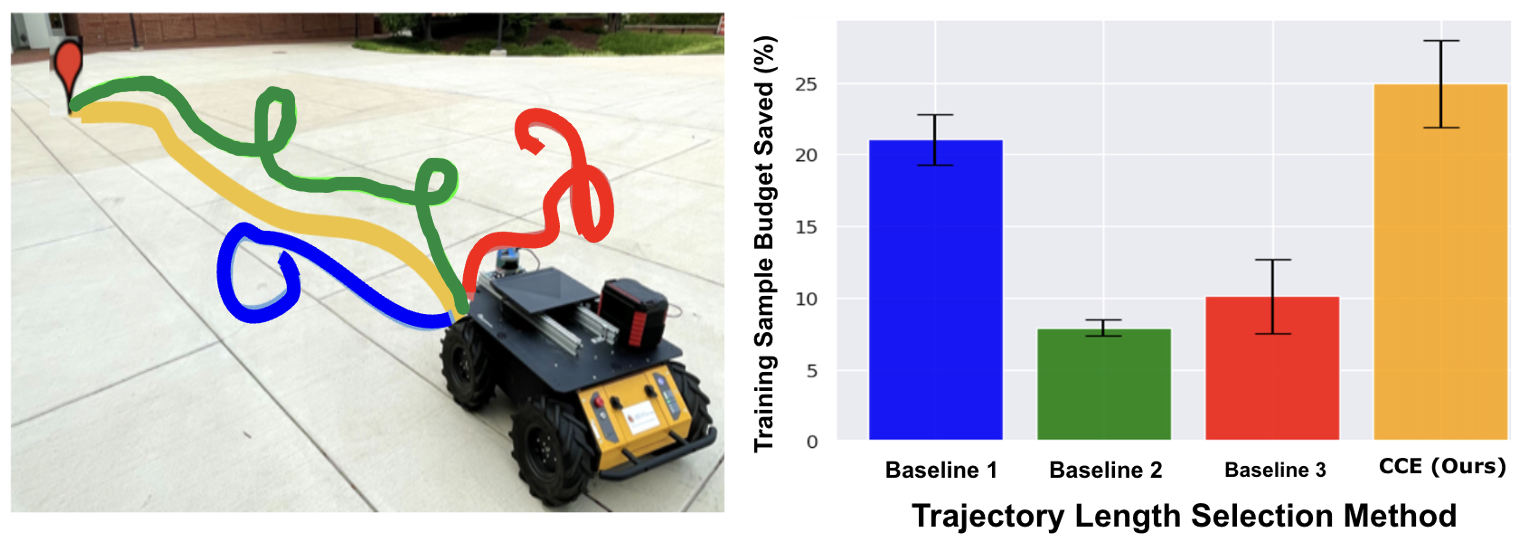}
  \caption{\small{The left figure shows sample trajectories of a robotic agent in a real-world test environment. The agent utilizes navigation policies trained with a policy gradient algorithm (REINFORCE) using four trajectory length schemes. We compare three baseline trajectory length schemes against \textbf{our CCE method (orange)}. CCE results in policies that generate shorter and more successful test-time trajectories while saving a larger percentage of the training sample budget as indicated in the right figure. More details on the baselines, policy gradient algorithm, and calculating percentage of sample budget saved can be found in Sec. \ref{sec:result}.}}
  \label{fig:teaser}
  \vspace{-7mm}
\end{figure}

To address this gap, we propose Confidence-Controlled Exploration (CCE), a novel strategy that balances exploration and exploitation without altering the reward structure. Instead of adding auxiliary rewards or fixed exploration bonuses, CCE leverages the agent’s policy entropy as a measure of confidence to guide exploration. The key idea is to dynamically balance the agent’s exploration and exploitation based on its uncertainty: when the policy is uncertain (high entropy), CCE encourages more exploration (e.g., by allowing the robot to sample longer trajectories before an update); when the policy is confident (low entropy), CCE restrains exploration to focus on exploiting the learned policy. In this way, the robotic agent explores only when needed, improving sample efficiency while still optimizing the original reward objective. Notably, this approach requires minimal modifications to existing RL algorithms, CCE can be applied on top of standard on-policy or off-policy methods by simply tuning a training hyperparameter (the trajectory length) based on policy entropy. We also highlight an interesting theoretical connection that our approach is utilizing between policy entropy and the policy’s mixing time in Markov processes \cite{levin2017markov, mutti2020intrinsically}, which provides principled guidance on how long an agent should explore to obtain reliable gradient estimates. In summary, our contributions are as follows:

\begin{itemize}
    \item \textbf{Confidence-Controlled Exploration (CCE):} We introduce CCE, a novel exploration mechanism for sparse-reward RL that uses policy entropy (an indicator of policy confidence) to balance exploration and exploitation in robotic navigation tasks.

    \item \textbf{No Reward Modification \& Enhanced Sample Efficiency:} CCE improves training sample efficiency by adaptively adjusting the agent’s trajectory length based on its confidence, without modifying the reward function or adding extrinsic bonuses. This robust exploration method preserves the original task objective.

    \item 	\textbf{Generality and Empirical Effectiveness:} CCE is a general plug-in approach that can integrate with different RL algorithms (demonstrated with both on-policy and off-policy methods) with minimal changes. We evaluate the navigation performance of CCE in both
simulated and real-world outdoor environments using a
Clearpath Husky robot. We observe that, for a fixed,
limited sample budget, CCE leads to an 18\% increase
in navigation success rate, a 20-38\% decrease in the
navigation path length, and a 9.32\% decrease in the
elevation cost compared to the policies from baselines.
\end{itemize}

%% file: sections/Related_Work.tex
\subsection{Related Works}







Motion planning and navigation have been well studied in robotics ~\cite{canny1988complexity,lavalle2006planning}. RL-based methods have been extensively used for robot navigations~\cite{motion_planning_survey}. Methods such as in \cite{weerakoon2021terp} used DRL to find reliable paths for outdoor navigation, while \cite{brunner2018teaching} learned a global map to find the shortest path out of a maze with RL. In \cite{staroverov2020real}, classical methods were used to create a hierarchical RL approach. For model-free RL, reward sparsity greatly affects policy search \cite{zai2020sparse_reward_definition, pathak2017curiosity} as a mobile robot will rarely receive any feedback to adjust its behavior. To produce more rewards that the algorithm can learn from, reward-shaping methods based on intrinsic rewards have been proposed \cite{ao2021}.
 
Methods in \cite{botteghi2020reward_shaping} and \cite{houthooft2016vime_reward_shaping} modified the reward output to encourage exploration of the environment. Furthermore, \cite{weerakoon2022htron} formulated a heavy-tailed policy to encourage exploration in sparse environments. Prior work used imitation learning and demonstration \cite{nair2018overcoming_demonstrations} to overcome the challenges of sparse rewards.  To balance exploration and exploitation, prior work has used the uncertainty of a surrogate cost function estimation from the visual field of view \cite{martinez2009bayesian} and a world model in model-based RL \cite{bhatia2022adaptive, yao2021sample}. In motion planning, \cite{rickert2008balancing} introduced a planner that balances exploration and exploitation using explicit information on the environmental information. Exploration-exploitation for multi-robot systems was considered in \cite{nakisa2014balancing, ghassemi2019decentralized, lee2021upper}. By connecting exploration to policy gradient estimation, our approach, CCE, is agnostic to data modality and does not require a world model or any explicit information on the given environment. 

%% file: sections/Problem_Formulation_condensed.tex
\section{Problem Formulation}


Consider a Markov Decision Process (MDP) characterized by the tuple $\mathcal{M} := (\mathcal{S}, \mathcal{A}, D, r)$, where $\mathcal{S}$ is the state space {for the robot in the environment}, $\mathcal{A}$ is the {set of possible actions the robot can take}, $D$ is the transition probability kernel that determines the next state $s' \in \mathcal{S}$ given $s \in \mathcal{S}, a \in \mathcal{A}$ via $s' \sim D(\cdot | s, a)$, and $r: \mathcal{S} \times \mathcal{A} \to [0, r_{\max}]$ is the reward function. The mobile robot navigates through this environment using a (potentially stochastic) policy $\pi$ that provides a conditional probability distribution over actions $a$ given a state $s$. Given a policy $\pi$ and transition kernel $D$, the transition matrix of the Markov chain induced by $\pi$ over $\mathcal{M}$ is $\mathbb P_\pi(s'|s) = \sum_{a \in \mathcal{A}} D(s'|s,a) \pi(a|s)$.
%
The goal in the RL setting is to find a policy maximizing long-run reward over $\mathcal{M}$. To achieve this, let $\mu$ be the unique stationary distribution to the Markov Chain. Thus, long-run reward is given by $J(\pi) = \mathbb{E}_{s \sim \mu, a \sim \pi} \left[ \sum_{t=0}^{\infty}  r(s_t, a_t) \right]$.
%
%
%
The expectation is over the policy and initial state distributions. In the average-reward settings, the state-action value function corresponding to a given policy $\pi$ is defined by $Q_{\pi}(s, a) = \mathbb{E}_{\pi} \left[ \sum_{t = 0}^{\infty}  r(s_t, a_t) - J(\pi) \ | \ s_0 = s, a_0 = a \right]$.

\vspace{2mm}
\noindent \textbf{Policy Gradient Algorithm.} To handle large state and action spaces, we focus on the case where policy $\pi$ is parameterized by a vector $\theta \in \mathbb R^d$, where $d$ denotes the parameter dimension, leading to the notion of a parameterized policy $\pi_{\theta}$. We define $J(\theta) = J(\pi_{\theta})$ and then, following policy gradient theorem \cite{sutton1999policy}, we note that   
%
\begin{align}\label{eqn:pgt}
   & \nabla J(\theta) = \mathbb{E}_{\pi_{\theta}, \mu_\theta} \left[ Q_{\pi_\theta}(s,a) \nabla \log \pi_{\theta}(a | s) \right],
\end{align}
which is the gradient of  $J(\theta)$. At update $u$ the policy gradient estimate is obtained by generating a $K \in \mathbb{N}$ length trajectory $\mathcal E_u:=\{s_i, a_i\}_{i=0}^{K-1}$ using $\pi_{\theta_u}$, and then computing
\begin{equation} \label{eqn:reinforce_pg}
    \widehat{ \nabla J(\theta_u) } =\frac{1}{K-1} \sum_{k=0}^{K-1} r(s_{k}, a_{k}) \nabla \log \pi_{\theta_u}(a_k | s_k).
\end{equation}
%
This expression is then used to perform a gradient update of the form $\theta_{u+1} = \theta_u + \eta \widehat{ \nabla J^(\theta_u) }$, for some stepsize $\eta > 0$. Each training sample $(s_k, a_k)$ is a simulation step. 

\vspace{2mm}
\noindent\textbf{Challenges due to sparse rewards.} 
In sparse reward settings, $r(s, a)$ in \eqref{eqn:reinforce_pg} is significantly invariant to different $(s, a)$ pairs. For example, in goal-reaching tasks where a positive reward is only granted when the goal is reached, the reward $r$ is $0$ for most steps in a trajectory $\mathcal E_u$. This reward structure makes it difficult to determine which action contributed to the goal if reached at all, significantly slowing down learning and even converging to suboptimal behaviors. Hence, exploration (visiting different parts of the state-action space ($\mathcal S \times \mathcal A$) to collect high rewards) is crucial to achieving reward-gaining trajectories. Prior methods to improve exploration \cite{pathak2017curiosity, MATARIC1994181, nair2018overcoming_demonstrations, chiang2019learning_reward_shaping, haarnoja2018soft} have high sample complexity from over-exploring, and they end up changing the reward and hence the learning objective, leading to learning instability \cite{sparse_reward_souradip}. Our goal in this work is to control the amount of exploration so that policy learning from gradient updates is sample-efficient and leads to stable performance. The challenges in controlling exploration throughout the training process lie in knowing how much exploration a given policy performs and how to implement the control without modifying the reward objective. To address these challenges, we look at the key concept of \textit{mixing time} that connects the amount of exploration of the agent to the gradient estimation.

\vspace{2mm}
\noindent \textbf{Mixing time and gradient estimation.}  In \eqref{eqn:pgt}, the expectation is taken with respect to the stationary distribution, $\mu_\theta$, of the transition matrix $\mathbb P_\theta$ induced by $\pi_{\theta}$. Intuitively, the mixing time \cite{suttlemixing2023} characterizes how many steps under $\pi_{\theta}$ are required before samples reflect $\mu_{\theta}$, which is critical for accurately estimating the policy gradient in \eqref{eqn:pgt}.  
When the mixing time is large, the trajectory length $K$ in \eqref{eqn:reinforce_pg} (which is the stochastic approximation of the gradient in \eqref{eqn:pgt}) must be sufficiently long to collect enough data for reliable gradient estimates \cite{dorfman2022, suttlemixing2023}. Conversely, when the mixing time is small, each environment step is more informative, allowing shorter trajectories to improve efficiency by freeing resources for more frequent policy updates. Mixing time is also closely related to the exploration of the state-space $\mathcal{S}$ \cite{dorfman2022, levin2017markov, riemer2022continual}, where low mixing time means more exploration. Therefore, adjusting the trajectory length $K$ in \eqref{eqn:reinforce_pg} based on mixing time effectively balances exploration and exploitation without altering the underlying reward $r$.
Unfortunately, mixing times are difficult to estimate in practice \cite{hsu2015, wolfermixing2020}, and there is no closed-form solution for calculating them. Thus, a suitable proxy is required for complex navigation tasks. In the next section, we demonstrate how leveraging the parameterized policy entropy provides a practical stand-in for mixing time, enabling control of exploration. 

%% file: sections/Proposed_Approach_condensed.tex
\section{Proposed Approach: Confidence Controlled Exploration-based Policy Learning} \label{sec:mapping}
\subsection{Policy Entropy to Control Navigation Exploration}

Prior works in stochastic optimization and RL have drawn ties between mixing time and the policy entropy \cite{mutti2020intrinsically, tarbouriech2019active}. These works have shown there exists a general trend where the policy entropy and the mixing time are inversely proportional. When a policy is more confident and deterministic, the exploration of the agent is lower and the mixing time is higher. Assuming a multivariate Gaussian policy, $\pi_{\theta}$, we define its entropy $H(\pi_{\theta})$ as the differential entropy
\begin{equation}\label{eq:gauss_entropy}
H(\pi_{\theta}) = - \frac{1}{|\mathcal{S}|}\sum_{s \in \mathcal{S}}\frac{1}{2}\log(( 2 \pi e)^N \det(\Sigma_{\theta}(s))),
\end{equation}
where $N$ is the action space dimension and $\Sigma_{\theta}(s)$ is the covariance matrix of the distribution $\pi_{\theta}(\cdot | s)$. 
Typically when using a neural network to train a policy $\pi_\theta$ for a continuous action space, the policy network outputs the estimated mean and covariance matrix $\Sigma_\theta$ of the action probability distribution for each $s$ inputted. Therefore, policy entropy with \eqref{eq:gauss_entropy} is simple to calculate repeatedly throughout training.

\vspace{1mm}
%
\noindent \textbf{Key idea: Policy entropy is a proxy for mixing time.} Ideally, the mixing time $\tau_{mix}$ should be used to determine trajectory length $K$. However, computing mixing time requires access to $\mathbb{P}_{\theta}$, which is impractical or impossible to accurately estimate in complex environments. Fortunately, policy entropy can be used a proxy for mixing time when selecting trajectory length $K$. To see why, notice that, assuming that the observed trend of policy entropy and mixing time being inversely proportional holds, we expect that a high-entropy or ``uncertain'' policy that performs significant exploration will have a shorter mixing time than a low-entropy, ``confident'' policy that performs little exploration. For this reason, uncertain, highly exploratory policies should have a shorter trajectory length $K$ than confident, minimally exploratory policies. Based on this reasoning, in this paper we use policy entropy \eqref{eq:gauss_entropy} as a practical proxy for mixing time that allows us to dynamically adjust $K$ throughout training.
We note that, while previous works have used policy entropy as a regularizer for the reward function to encourage exploration in the sparse reward setting \cite{haarnoja2018soft, eysenbachmaximum}, we propose using policy entropy to control exploration by dynamically adjusting trajectory length $K$ without changing the reward objective $r$. To the best of our knowledge, we are the first to apply the observed inverse proportionality between mixing time and policy entropy to augment the PG training protocol of robotic navigation agents.

\subsection{CCE: Confidence-Controlled Exploration}


The key to our approach is to select trajectory length based on the current value of the policy entropy, our measure for confidence. We perform this computation using a monotonic mapping $f$ from the range of possible policy entropy values to a range of possible trajectory length values. In this paper, we utilize a linear function, but we note that this is a design choice and that $f$ can be implemented with any monotonically decreasing mapping of policy entropy to $K$. Let $H_0$ denote the initial policy entropy value and $H_c$ the entropy of the current policy. In practice, we choose $H_0 = H(\pi_{\theta_0})$ to be the entropy of the initial policy, since policy entropy tends to decrease during training. Let $K_0 \in \mathbb{N}$ denote the user-specified initial trajectory lengths and let $\alpha \geq K_0$ denote a user-specified sensitivity parameter that controls the rate of change of trajectory length. For our experiments, we determine the trajectory length, $K_c$, corresponding to $H_c$ using the mapping $f$ below:
%
\begin{equation}\label{eq:mapping}
 K_c = f^{H_0}_{ K_0, \alpha}(H_c) = \text{round}  \left( \alpha - \frac{H_c}{H_0}(\alpha - K_0)\right),
\end{equation}
%
where the rounding operation is to ensure that $K_c$ is always an integer. When $H_c = H_0$, we have the initial $K_0$. Theoretically, if $H_c$ ever reached $0$, we would have $K_c = \alpha$. Because in practice $H_c$ would not reach $0$, $\alpha$ is much larger than the ending trajectory lengths in our experiments. 
Furthermore, in our experiments, we initialize our policies with a Gaussian distribution. If needed, to ensure that $H_c$ never increases too much above $H_0$ so that $K_c$ never becomes $0$, one could initialize with a uniform distribution, so that $H_0$ is the maximum possible entropy. Algorithm \ref{algo:one} summarizes our CCE methodology for adaptive trajectory length PG methods. Note that any on-policy PG method can carry out the policy gradient update in Line \ref{line:pg}.
The training is divided into episodes where the start location and goal are constant between episodes. During each episode, a gradient update occurs after every trajectory. The episode ends when the goal is reached or when a user-set number $T$ of simulation steps has occurred. Given an episode, the trajectory length is fixed. $K_c$ is only adjusted based on the policy entropy at the start of the new episode.

\begin{algorithm}[]
	\begin{algorithmic}[1]
		\STATE \textbf{Initialize:} policy parameter $\theta_0$, policy entropy $H_0$, trajectory length $K_0$, current trajectory length $K_c = K_0$, number of episodes $E$, maximum number $T$ of sample/steps per episode, sensitivity parameter $\alpha$, update index $u = 0$ \\
        \FOR{$\text{episode} = 0, \ldots, E-1$}
            \STATE calculate $H_c$ using \ref{eq:gauss_entropy} 
            \STATE calculate $K_c$ using \ref{eq:mapping}
            \STATE $t = 0$
    		\WHILE{$t < T$}
                \STATE $\mathcal E_u = \{\};\ k = 0$
                \WHILE{$k < K_c$ \textbf{and} $t < T$}
        	\STATE collect $(s_k, a_k, r_k)$ with $\pi_{\theta_u}$ and add to $\mathcal E_u$ \\ 
                \STATE $k = k + 1;\ t = t + 1$
                \ENDWHILE
                \STATE \textbf{Perform} gradient update using $\mathcal{E}_u$ to obtain $\theta_{u+1}$ \label{line:pg}
                
                \STATE $u = u + 1$
            \ENDWHILE
        \ENDFOR
            
		

		\STATE \textbf{Return:} $\pi_{\theta_u}$ \\ 
	\end{algorithmic}

	\caption{Policy Learning with CCE}
	\label{algo:one}

\end{algorithm}

%% file: sections/Experiments.tex
\section{Experiments and Results}
\label{sec:result}

\begin{figure}[t] 
    \centering  \includegraphics[width=\columnwidth]{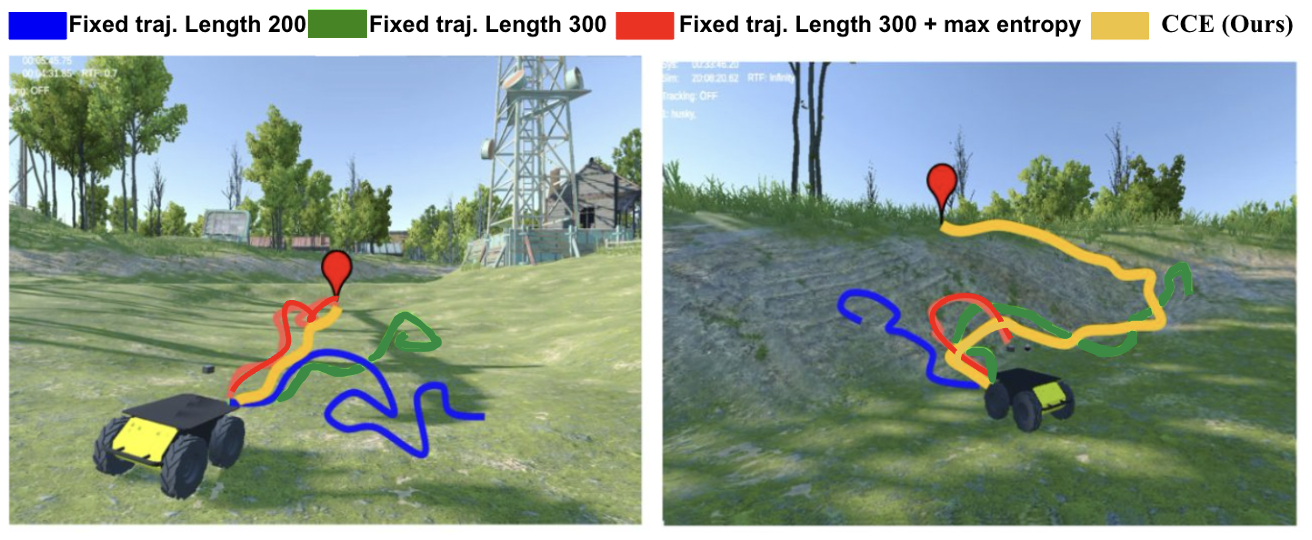}
    \caption{\small{Comparison of four different trajectory-length schemes on \small{\textbf{[LEFT]} even and \textbf{[RIGHT]} uneven terrain navigation tasks in Unity-based outdoor robot simulator. For REINFORCE, PPO, and SAC, using CCE as the trajectory length scheme for $150$ episodes leads to a policy with a shorter path length return than baselines with limited samples. The drawn trajectories are sample representations of the robot’s odometry data collected.
    %
    %
    The trajectories pictured are representative of all $3$ algorithms since we found that each trajectory-length scheme leads to similar navigation regardless of the RL algorithm.
    See Figures \ref{fig:even_training} and \ref{fig:uneven_training} for training details.}}}
    \label{fig:terrain_nav}
    \vspace{-12pt}
\end{figure}

\begin{figure*}[ht] 
    \centering  \includegraphics[width=0.95\textwidth]{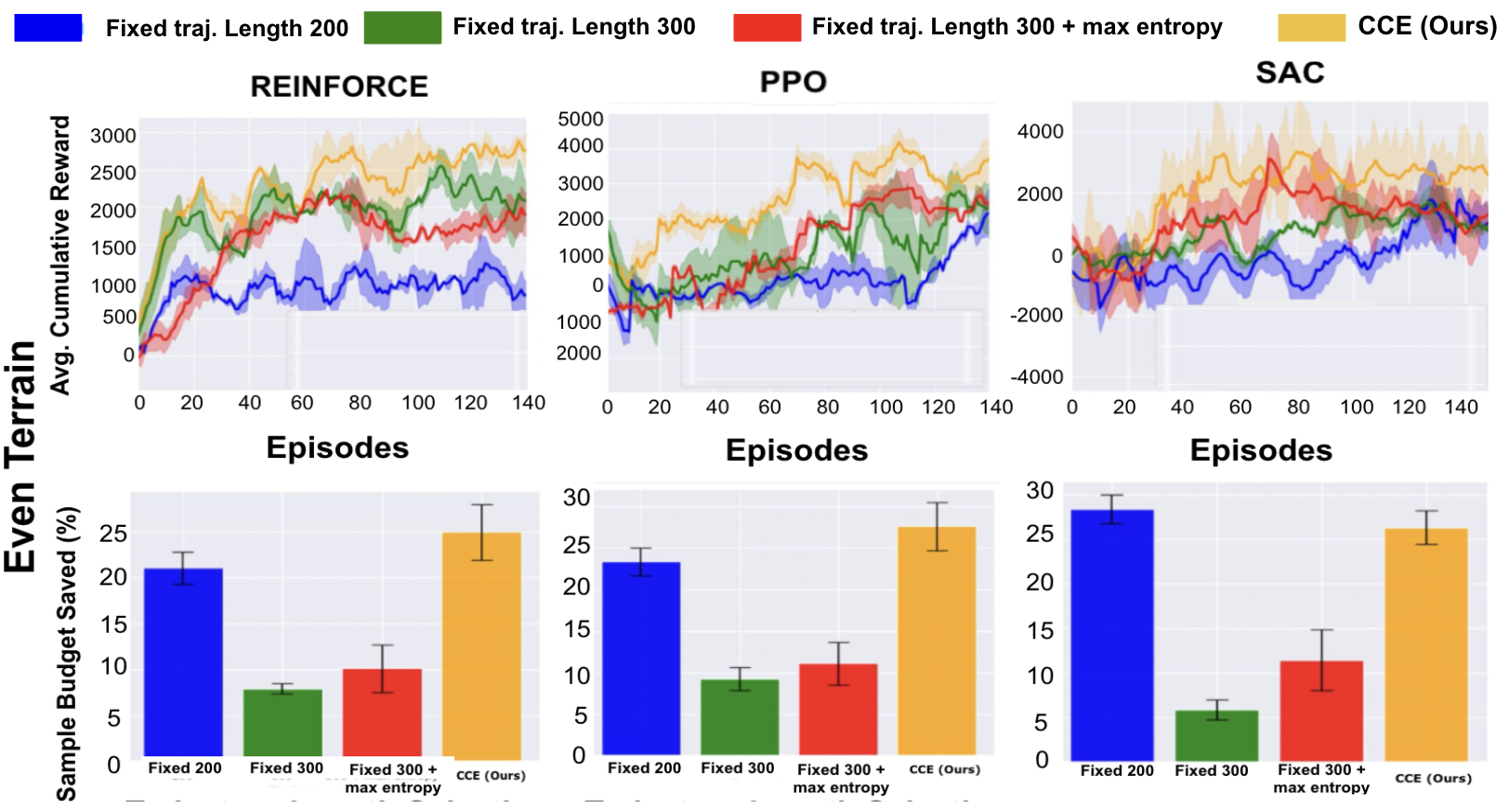}
    %
    \caption{\small{
    \textbf{[TOP ROW]} Learning curves during training and \textbf{[BOTTOM]} percent of sample budget ($\sim6.0e4$) saved during navigation tasks for the even terrain (left of Figure \ref{fig:terrain_nav}).
    Comparison of \textbf{[LEFT COLUMN]} REINFORCE, \textbf{[MIDDLE]} PPO, and \textbf{[RIGHT]} SAC, with constant and adaptive trajectory lengths.
    For each algorithm, CCE converges to a higher cumulative reward while expending less of the total budget. We ran $8$ to $10$ independent replications for each of the $12$ experiments. Although a fixed trajectory length of $200$ saves the sample budget comparably to CCE, it achieves lower reward in all cases due to the robot flipping over thus causing episodes to end.  
    }}
    \label{fig:even_training}
\end{figure*}

\begin{figure*}[t] 
    \centering   \includegraphics[width=0.95\textwidth]{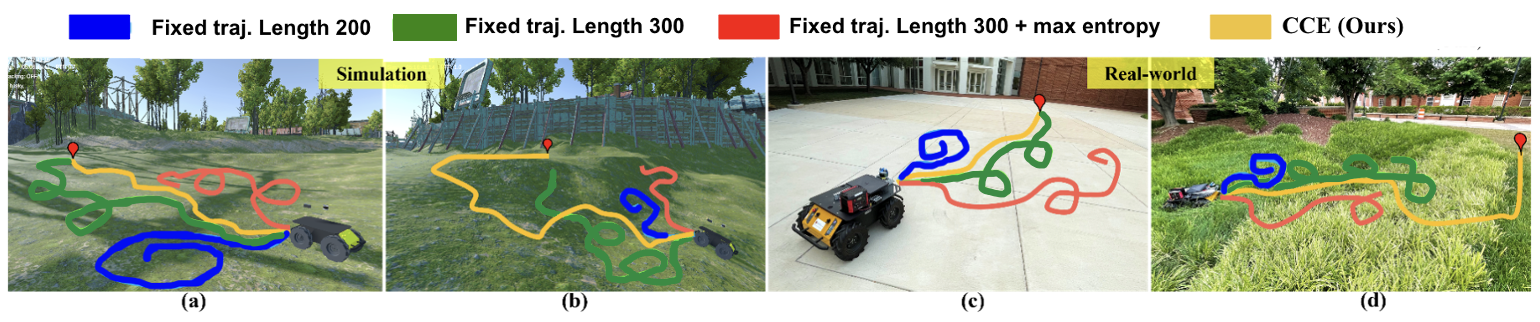}
    \caption{\small{
    %
    Sample navigation trajectories generated in \textbf{[TOP ROW]} simulated  and \textbf{[BOTTOM]} real-world environments by policies trained with REINFORCE using CCE vs. fixed trajectory lengths at a fixed sample budget of $\sim 4.5e4$. We observed that the fixed trajectory length policies often fail at test time due to inefficient use of the available sample budget during training. In contrast, CCE policies successfully complete the tasks after training with the same limited sample budget. The experiments with a real Clearpath Husky robot validate that CCE can be transferred onto real robotic systems without significant performance degradation. The drawn trajectories are sample representations of the robot's odometry data. 
    }}
    \label{fig:nav_comparisons}
\end{figure*}

%
%
%
\textbf{RL Algorithms.} We choose three RL algorithms to integrate CCE into for improved sample efficiency: REINFORCE, PPO, and SAC. As REINFORCE and PPO are both on-policy policy gradient methods and SAC is an off-policy method, we show that CCE can improve the sample efficiency of various classes of policy gradient algorithms. For PPO and SAC, we adaptively change the size of the replay buffer. We use feedforward neural networks with a set of linear layers followed by ReLu activation as the backbone in all the RL algorithms. The actor and the value functions use the same network backbone that includes linear layers of size $64$, $128$, $512$, $128$, and $64$ in that order. The input and output layer sizes vary based on the shape of the state input and the desired output. For more information on hyperparameters, please see the Appendix of our supplemental version.
%
%
%

\textbf{Baselines.} Since our proposed approach relies on adaptive trajectory length, for each algorithm we test CCE with, we compare it against the same algorithm with fixed trajectory lengths. Because in all our robotic navigation experiments, CCE-based training schemes have trajectory lengths start at $180-200$ and end at $300$, we set baseline fixed trajectory lengths at $200$ and $300$ to compare against policies that were trained with trajectory lengths on the lower and higher end of the CCE trajectory length distributions.
We also compare with a fixed trajectory length of $300$ with max-entropy regularizer as that encourages exploration. 
 We also incorporate a classical navigation algorithm Ego-graph \cite{ego_graph2}, to highlight the navigation performance of our method under challenging terrain conditions. Ego-graph is a baseline method in \cite{ego_graph2} that utilizes elevation data to obtain actions that minimize elevation gradient cost.
 
 \textbf{Environments.} We utilize two environments in a unity-based outdoor robot simulator. Finally, we deploy navigation policies trained for a fixed sample budget using the robot simulator on a real Clearpath Husky robot. We present the details of the training environments below:
%
%
\begin{figure*}[ht] 
    \centering  \includegraphics[width=0.95\textwidth]{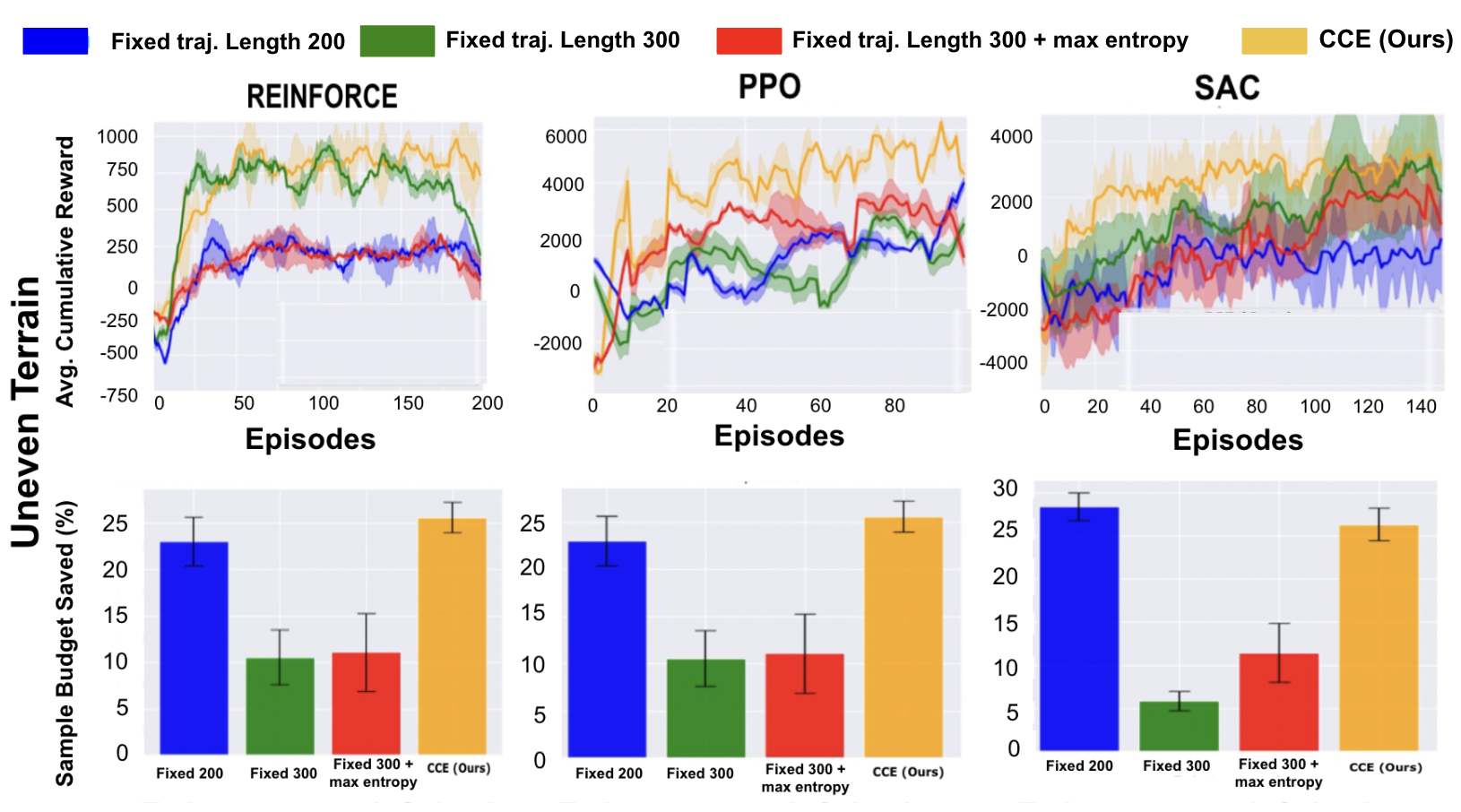}
    \caption{\small{
    \textbf{[TOP ROW]} Learning curves during training and \textbf{[BOTTOM]} percent of sample budget ($\sim6.0e4$) saved during navigation tasks for the uneven terrain (right of Figure \ref{fig:terrain_nav}).
    Comparison of \textbf{[LEFT COLUMN]} REINFORCE, \textbf{[MIDDLE]} PPO, and \textbf{[RIGHT]} SAC, with constant and adaptive trajectory lengths.
    For each algorithm, CCE converges to a higher cumulative reward while expending less of the total budget. We ran $8$ to $10$ independent replications for each of the $12$ experiments. Although a fixed trajectory length of $200$ saves the sample budget comparably to CCE, it achieves lower reward in all cases due to the robot flipping over thus causing episodes to end.
    }}
    \label{fig:uneven_training}
\end{figure*}


\noindent \textbf{Simulated Even and Uneven Terrain Navigation.} We use a Unity-based outdoor simulator with a Clearpath Husky robot model to train policies for two navigation tasks: 1) Goal Reaching and 2) Uneven terrain navigation. The Unity simulator includes diverse terrains and elevations for training and testing. We restrict ourselves to a $100m \times 100m$ region during training. 
We used identical policy distribution parameters and neural network architectures with all four algorithms during training and evaluation. The policies are trained using Pytorch in the simulator equipped with a Clearpath Husky robot model, and a ROS Melodic platform. The training simulations are executed on a workstation with an Intel Xeon 3.6 GHz CPU and an Nvidia Titan GPU. 
    
Our agent is a differential drive robot with a two-dimensional continuous action space $a = (v,\omega)$ (i.e. linear and angular velocities). The simulator includes even and uneven outdoor terrains and the state space varies for the two navigation tasks and is obtained in real-time from the simulated robot's odometry and IMU sensors. The state inputs used in the two scenarios are

\begin{itemize}
\item $d_{goal} \in [0,1]$- Normalized distance (w.r.t the initial straight-line distance) between the robot and its goal;
\item  $\zeta_{goal} \in [0,\pi]$- Angle between the robot’s heading direction and the goal;
\item $(\phi,\theta) \in [0,\pi]$- Roll and pitch of robot, respectively.
\end{itemize}

Inspired by \cite{weerakoon2022htron}, we define three reward functions to formulate the sparse rewards to train our navigation policies: 1) Goal distance reward $r_{dist}$, 2) Goal heading reward $r_{head}$, and 3) Elevation reward $r_{elev}$ to formulate the sparse rewards to train our navigation policies. Hence $ r_{dist} = \frac{\beta}{2}\mathcal{N}(\frac{d_{goal}}{2},\sigma^{2}) + \beta\mathcal{N}(0,\sigma^{2})$ and $r_{head} = \mathbb{I}_{ \{|\zeta_{goal}| \leq \pi/3 \}}$ 
%
%
where $\mathcal{N}$ denotes Gaussian distribution with variance $\sigma$ and $\beta$ is the amplitude parameter. We set the variance $\sigma = 0.2$ to ensure that the $r_{dist}$ only includes two narrow reward peaks near the goal and half-way from the goal. Similarly, $r_{elev}$ is defined as $r_{elev} = \mathbb{I}_{ \{|\phi| \geq \pi/6 \}} \cup \mathbb{I}_{ \{|\theta| \geq \pi/6\}}$.
Our reward definitions are significantly sparse compared to the traditional dense rewards used in the literature \cite{weerakoon2021terp}. Below are the three main types of experiments:

\textbf{(a) Even terrain navigation.} The robot is placed in an obstacle-free even terrain environment and random goals in the range of 10-15 meters away from the starting position given at each episode. The state space is $s = [d_{goal},\zeta_{goal}]$ and the overall reward formulation $r_{even} = r_{dist} + r_{head}$. 

\textbf{(b) Uneven terrain navigation.} The robot is placed in an obstacle-free, highly uneven terrain environment. To account for this complexity, we input the roll and pitch of the angle of the robot into the state space, $s = [d_{goal},\zeta_{goal},\phi,\theta]$. The objective is to navigate the robot to a goal location along relatively even terrain regions to avoid possible robot flip-overs. The reward is $r_{uneven} = r_{dist} + r_{head} - r_{elev}$. 

\textbf{(c) Real-world Experiments.} The real and simulated testing were conducted in different outdoor scenarios. Particularly, the real-world environments are reasonably different in terms of terrain structure (i.e. elevation gradient) and surface properties (i.e grass and tiny gravel regions with different levels of friction) which are not included or modeled in the simulator (e.g. elevation gain in the simulator is up to $\sim 4m$, however, we restricted it to $\sim 1.5m$ elevation gain in the real world for robot’s safety). Our Clearpath Husky robot is equipped with a VLP16 LiDAR and a laptop with an Intel i9 CPU and an Nvidia RTX 2080 GPU.

\AtBeginEnvironment{tabular}{} 
\begin{table*}[h!]

\resizebox{\textwidth}{!}{
\begin{tabular}{|c| c |cc| cc| cc|} 
\hline

\textbf{Terrain}  & \textbf{Trajectory Length Method} & \multicolumn{2}{c|}{\textbf{SR}}  & \multicolumn{2}{c|}{\textbf{PL}}  & \multicolumn{2}{c|}{\textbf{EC}}  \\ [0.5ex] 

 & \textbf{used with REINFORCE Algorithm} &  \multicolumn{2}{c|}{(\%)$\uparrow$} &  \multicolumn{2}{c|}{(\# m)$\downarrow$}  &  \multicolumn{2}{c|}{(m)$\downarrow$}   \\ [1ex]\hline 
 & & Simulation & Real World & Simulation & Real World & Simulation & Real World \\ \hline
\textbf{Even}
& Traj. length = 300 + max entropy  & 58 & 40 & 22.47 &  \textbf{9.32}& - & -  \\
& Traj. length = 200              & 32 &  30& 25.58 & 15.83& - & - \\
& Traj. length = 300 & 65 & 50 & 18.75 & 10.25& -  & - \\
& CCE (Ours)              & \textbf{84} & \textbf{80} & \textbf{11.62} &  9.66& - & -  \\
\hline

\textbf{Uneven}
& Ego-graph \cite{ego_graph2}  & 54 & 60 & 20.36 & \textbf{15.29}& 2.189 & \textbf{0.758} \\
& Traj. length = 300 + max entropy            & 28 & 20 & 41.32 & 20.67& 3.163 & 1.026\\
& Traj. length = 200  & 8 & 10 & \textbf{9.25} & 22.08& 3.116 &  1.113\\
& Traj. length = 300 & 31 & 20 & 32.44 & 18.94& 2.236 &   0.997\\
& CCE (Ours)               & \textbf{72} & \textbf{70} & 16.28 & 16.33& \textbf{1.985}  & 0.820\\
\hline
\end{tabular}
}
\caption{\small{
\textbf{Robotic navigation performance} of the policies trained for even and uneven terrain navigation tasks with limited samples. SR denotes success rate, PL denotes average path length, and EC denotes elevation cost. Each method is tested for 100 trials in the outdoor simulator and for 10 trials in real-world environments. Test environments are different from training environments. The average values are reported in the table. CCE has a higher SR across both terrains in simulation and real-world terrains. For most cases where it does not achieve the lowest PL or EC, CCE has the second lowest which is usually comparable to the method with the lowest PL or EC.
}}
\label{tab:nav_results}
\vspace{-6pt}
\end{table*}

\begin{figure*}[ht] 
\centering
    \includegraphics[width=0.9\textwidth]{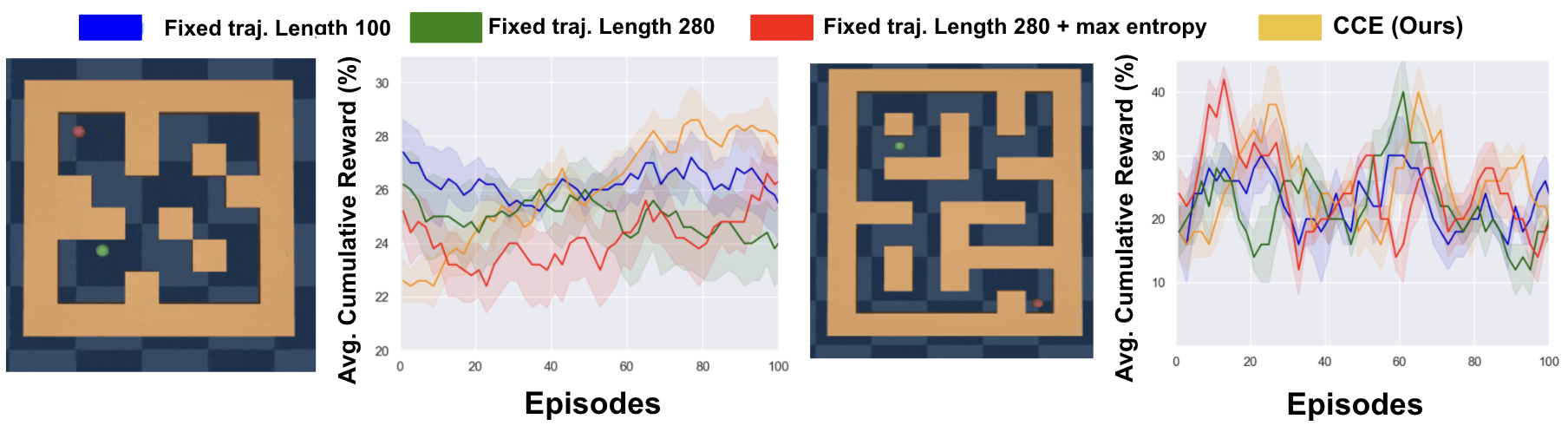}
    \caption{\small{Learning curves during training in the MuJoCo \cite{mujoco} Maze environments, where the green ball tries to navigate towards the red ball. We utilize our CCE (orange) approach against the baselines). In the leftmost two figures, we see that CCE attains higher training performance than other trajectory selection methods in PointMaze-Medium. However, in the PointMaze-Large maze environment on the right, we see that CCE performs comparably to other baseline methods. We believe this similar performance is due to the number of bottlenecks, reducing the number of paths between states, weakening the connection between mixing time and policy entropy. 
    %
}}\label{fig:pointmaze}
\end{figure*}

 


\subsection{Evaluation Metrics}

\textbf{PG Simulation Training.} \textit{1) Training Performance:} Average cumulative rewards across episodes. \textit{2) Sample Efficiency:} The percentage of training sample budget saved, $1 - p_{\text{used}}$, where $p_{\text{used}}$ is the number of total samples used by the given PG algorithm divided by the total sample budget, $6.0e4$ ($150$ episodes with max samples of $400$ each).

\textbf{Quality of Test-time Navigation Path Generated.}\textit{ 1) Success Rate (SR):} The percentage of successful goal-reaching attempts without robot flip-overs (especially in uneven terrain) out of the total number of trials. \textit{2) Average path length (PL):} The average length of the path taken to reach the goal 10 meters away from the starting position. \textit{3)Elevation cost (EC):} The total elevation gradient experienced by the robot during a trial (i.e., if $z_{r}$ is the vector that includes the gradient of the vertical motions of the robot along a path, the elevation cost is given by $||\nabla z_{r}||$).



%

\subsection{Results and Discussions}
%


\textbf{Policy Learning.} 
We observe in Figure \ref{fig:even_training} and \ref{fig:uneven_training} that, for the outdoor robot simulator, CCE achieves similar or better reward returns with significantly fewer samples under both the navigation settings compared to the fixed trajectory lengths and entropy regularization. Specifically, despite providing comparable returns to CCE, the sample efficiency of the constant and large trajectory lengths is significantly lower (see the bar plots in Fig. \ref{fig:even_training} and \ref{fig:uneven_training}). 
For the agents using CCE, in the even terrain simulations, all agents started with an initial trajectory length of $180$, and in the uneven terrain, $200$. All those agents end with a final length of $300$. 


\textbf{Robotic Navigation Performance.} We compare our method’s navigation performance qualitatively in Fig. \ref{fig:nav_comparisons} and quantitatively in Table \ref{tab:nav_results}. To highlight the importance of sample efficiency during training, we use policies trained up to episode $150$ with the \textbf{\textit{limited}} sample budget of $6.0e4$ to perform navigation tasks in even and uneven simulated terrains. The policies, except for ones from CCE, cannot consistently complete the navigation tasks, resulting in significantly lower success rates. Even successful trials of these policies lead to significantly longer paths with several loops instead of heading toward the goal. In contrast, CCE results in relatively straight and consistent paths toward the goals. We further observe that uneven terrain navigation is a comparatively challenging task even for traditional methods such as Ego-graph\cite{ego_graph2}. Hence, the success rates are relatively low due to the robot flip-overs that occurred during trials. However, CCE leads to higher success rates even compared to baseline methods such as Ego-graph. We further observe that CCE results in lower elevation cost indicating that the generated paths are smoother in uneven terrain conditions. Finally, we deploy the policies trained with the outdoor simulator in the real world using a Clearpath Husky robot. In Table \ref{tab:nav_results}, even and uneven terrain navigation policies with CCE can be transferred to the real world without significant performance degradation in success rate, path length, and elevation cost while other policies show inconsistent performance similar to the simulations.

\noindent \textbf{Remark.} Given more training samples, the baseline methods could perform better in success rate and path length during test time given that Figures \ref{fig:even_training} and \ref{fig:uneven_training} show the many times they can reach similar cumulative return as CCE. However, we tested policies from earlier in training to highlight that CCE does not need as many samples. 

\subsection{Performance in Environments with Bottlenecks}
To further test our CCE approach, we utilize MuJoCo simulation in the sparse reward PointMaze environments. Figure \ref{fig:pointmaze} shows the agent (green ball) navigating towards the goal (red ball) in two different maze environments: PointMaze-Medium maze (left) and PointMaze-Large maze (right). The agent gains a sparse reward of +1 for reaching within 0.45 units of goal and +0 otherwise. The episode terminates when the agent reaches the goal or the maximum number of steps is reached. Using REINFORCE, we train $4$ policies with CCE and baseline fixed trajectory lengths of $100$, $280$, and $280$ + max entropy regularizer. More details on the environments, policy network and hyperparameters can be found in the Appendix of the supplemental version. We run each experiment for 5 trials. CCE achieves better training performance in the PointMaze-Medium environment over baselines but shows similar performance in PointMaze-Large. A possible reason is that the correlation between mixing time and policy entropy is weak when the environment has many bottlenecks. Prior work shows that maximizing state visitation and exploration becomes increasingly difficult with bottlenecks, inducing a slow mixing $\mathbb P_\theta$, because of the fewer paths between states in $\mathcal S$ \cite{riemer2022continual, levin2017markov, montenegro2006mathematical}. In such environments, a high-entropy policy will not necessarily induce a fast-mixing $\mathbb P_\theta$, breaking the assumption that the general trend between mixing time and policy entropy holds. 

%% file: sections/Conclusion.tex
\section{Conclusions, Limitations, and Future Works}
\label{sec:conclusion}
We propose CCE, which controls exploration and induces training sample efficiency for goal-reaching navigation in the sparse reward setting with RL-based methods by adaptively changing the trajectory length during training. We do so by utilizing a theoretical connection between the entropy of the robotic agent's policy and the exploration of the agent. We show that CCE induces more sample-efficient training than baselines. On real robots, CCE-trained policies generally achieve better success rates and shorter path lengths than ones trained with constant trajectory lengths and entropy regularization for the same sample budget. In real-world experiments, policies often lead to high motor vibrations from infeasible velocities provided as actions. Hence, further investigation is required to generate dynamically feasible and smooth actions for better sim-to-real transfer. As shown in the MuJoCo experiments, one limitation is that CCE performs comparably to baselines when in environments with significant bottlenecks and fewer paths between states. Future work could analyze this effect in non-stationary environments or in other robotics applications, such as control tasks. 


%% file: sections/Appendix.tex
\section{Appendix}\label{section:appendix}


\subsection{Experimental Setup Details}

Here are the full descriptions of navigation experiments. The first one details the Clearpath Husky simulation experiments in even and uneven terrain shown in the main body. The second one explains the real robotic experiments also shown in the main body. The third set of experiments shows the benefits of CCE with Actor-Critic (AC) in 2D gridworld:
\begin{enumerate}


    \item \textbf{Even and Uneven Terrain Navigation in Robotic Simulations}: We use a Unity-based outdoor simulator with a Clearpath Husky robot model to train policies for two navigation tasks: 1. Goal Reaching, 2. Uneven terrain navigation using the Reinforce algorithms. The Unity simulator includes diverse terrains and elevations for training and testing. We restrict ourselves to a $100m \times 100m$ region during training. 
     We used identical policy distribution parameters and neural network architectures with all four algorithms during training and evaluation. The policies are implemented using Pytorch to train in the simulator equipped with a Clearpath Husky robot model, and a ROS Melodic platform. The training and simulations are executed on a workstation with an Intel Xeon 3.6 GHz CPU and an Nvidia Titan GPU. 
    
    Our agent is a differential drive robot with a two-dimensional continuous action space $a = (v,\omega)$ (i.e. linear and angular velocities). The state space varies for the two navigation tasks and is obtained in real time from the simulated robot's odometer and IMU sensors. The state inputs used in the two scenarios are,

    \begin{itemize}[]
    \item $d_{goal} \in [0,1]$ - Normalized distance (w.r.t the initial straight line distance) between the robot and its goal;
    \item  $\alpha_{goal} \in [0,\pi]$ - Angle between the robot’s heading direction and the goal;
    \item $(\phi,\theta) \in [0,\pi]$ - Roll and pitch angle of the robot respectively;
    \end{itemize}

    Inspired by \cite{weerakoon2022htron}, we define three reward functions: Goal distance reward $r_{dist}$, Goal heading reward $r_{head}$, and Elevation reward $r_{elev}$ to formulate the sparse rewards to train our navigation policies. Hence,

     \begin{equation}
     r_{dist} = \frac{\beta}{2}\mathcal{N}(\frac{d_{goal}}{2},\sigma^{2}) + \beta\mathcal{N}(0,\sigma^{2}) \ \ \text{and} \\, r_{head} = \mathbbm{1}_{ \{|\alpha_{goal}| \leq \pi/3 \}} \ \, 
    \end{equation}

    where, $\mathcal{N}$ denotes Gaussian distribution with variance $\sigma$ and $\beta$ is the amplitude parameter. We set the variance $\sigma = 0.2$ to ensure that the $r_{dist}$ only includes two narrow reward peaks near the goal and half-way from the goal. Similarly, $r_{elev}$ is defines as,

    \begin{equation}
    r_{elev} = \mathbbm{1}_{ \{|\phi| \geq \pi/6 \}} \cup \mathbbm{1}_{ \{|\theta| \geq \pi/6\}}.
    \label{eq:r_stable}
    \end{equation}

    We observe that our reward definitions are significantly sparse compared to the traditional dense rewards used in literature\cite{weerakoon2021terp}.
    
    \textbf{(a) Even terrain navigation:} The robot is placed in an obstacle-free even terrain environment and random goals in the range of 10-15 meters away from the starting position is given at each episode. The state space is $s = [d_{goal},\alpha_{goal}]$ and the overall reward formulation $r_{even} = r_{goal} + r_{head}$.

    \textbf{(b) Uneven terrain navigation:} The robot is placed in an obstacle-free, highly uneven terrain environment. The state space is $s = [d_{goal},\alpha_{goal},\phi,\theta]$. The objective is to navigate the robot to a goal location along relatively even terrain regions to avoid possible robot flip overs. The overall reward formulation $r_{uneven} = r_{goal} + r_{head} + r_{elev}$.

    \item \textbf{Real-world Experiments:} The real and simulated testing were conducted in different outdoor scenarios. Particularly, the real-world environments are reasonably different in terms of terrain structure(i.e. elevation gradient) and surface properties(i.e grass and tiny gravel regions with different levels of friction) which are not included or modeled in the simulator (e.g. elevation gain in the simulator is up to $\sim 4m$, however, we restricted it to $\sim 1.5m$ elevation gain in the real world for robot’s safety). Our Clearpath Husky robot is equipped with a VLP16 LiDAR and a laptop with an Intel i9 CPU and an Nvidia RTX 2080 GPU.

    \item \textbf{MuJoCo Maze Environments:}  The state space is a $4$-dimensional vector representing the agent's $(x,y)$ positional coordinates and the velocities in each direction. The action space is a $2$-dimensional vector representing the $(x,y)$ force actuated on the agent. Each element in the action vector is between $-1$ and $1$, inclusive. The agent gains a sparse reward of $+1$ for reaching within $0.45$ units of goal and +0 otherwise. The episode terminates when the agent reaches the goal or the maximum number of steps is reached. We use a Gaussian neural network with $2$ hidden layers, each of size $256$.

    
\end{enumerate}

\begin{table}[h!]
\caption{This table compares the hyperparameters and performance between the four experiments, each run for five trials}
\label{table:hyperparameters}
\begin{center}
\begin{tabular}{|c|c|c|c|c|c|c|}
\hline
Environment & Model & Trajectory Method & $\alpha$ & Trajectory Change & Learning Rate & No. of Iterations Per Episode \\
\hline

\textbf{Robot Simulations} & & & & & &\\
\hline
 & REINFORCE & Fixed = 200 &  &  & 0.0001  & 400\\
Even Terrain  & REINFORCE & Fixed = 300 &  &  & 0.0001  & 400\\
 & REINFORCE & CCE & 3200 & 180 $\to$ 300 & 0.0001  & 400\\
 & REINFORCE & Fixed = 300 + max entropy &  &  & 0.0001  & 400\\
\hline
 & REINFORCE & Fixed = 200 &  &  & 0.0001  & 400\\
Uneven Terrain  & REINFORCE & Fixed = 300 &  &  & 0.0001  & 400\\
 & REINFORCE & CCE & 3200 & 200 $\to$ 300 & 0.0001  & 400\\
 & REINFORCE & Fixed = 300 + max entropy   &  &  & 0.0001  & 400\\
\hline

\hline
 & PPO & Fixed = 200 &  &  & 0.0005  & 400\\
Even Terrain  & PPO  & Fixed = 300 &  &  & 0.0005  & 400\\
 & PPO  & CCE & 3200 & 180 $\to$ 300 & 0.0005  & 400\\
 & PPO  & Fixed = 300 + max entropy &  &  & 0.0005  & 400\\
\hline
 & PPO  & Fixed = 200 &  &  & 0.0005  & 400\\
Uneven Terrain  & PPO & Fixed = 300 &  &  & 0.0005  & 400\\
 & PPO  & CCE & 3200 & 200 $\to$ 300 & 0.0005  & 400\\
 & PPO  & Fixed = 300 + max entropy   &  &  & 0.0005  & 400\\

\hline
 & SAC & Fixed = 200 &  &  & 0.0003  & 400\\
Even Terrain  & SAC & Fixed = 300 &  &  & 0.0003  & 400\\
 & SAC & CCE & 3200 & 180 $\to$ 300 & 0.0003  & 400\\
 & SAC & Fixed = 300 + max entropy &  &  & 0.0003  & 400\\
\hline
 & SAC & Fixed = 200 &  &  & 0.0003  & 400\\
Uneven Terrain  & SAC & Fixed = 300 &  &  & 0.0003  & 400\\
 & SAC & CCE & 3200 & 200 $\to$ 300 & 0.0003  & 400\\
 & SAC & Fixed = 300 + max entropy   &  &  & 0.0003  & 400\\

 & & & & & &\\

\hline

\textbf{MuJoCo PointMaze} & & & & & &\\
\hline

Medium & REINFORCE & Fixed = 100 &  &  & 0.01  & 2000\\
Medium & REINFORCE & Fixed = 280 &  &  & 0.01  & 2000\\
Medium & REINFORCE & CCE & 50000 & 100 $\to$ 280 & 0.01  & 2000\\
Medium & REINFORCE + max-entropy & Fixed = 280 &  &  & 0.01  & 2000\\
\hline
Large & REINFORCE & Fixed = 100 &  &  & 0.01  & 10000\\
Large & REINFORCE & Fixed = 280 &  &  & 0.01  & 10000\\
Large & REINFORCE & CCE & 100000 & 100 $\to$ 280 & 0.01  & 10000\\
Large & REINFORCE + max-entropy & Fixed = 280 &  &  & 0.01  & 10000\\
\hline

\end{tabular}
\end{center}
\end{table}